# Generalizing the Dempster-Shafer Theory to Fuzzy Sets[1]

John Yen

USC / Information Sciences Institute
4676 Admiralty Way,
Marina del Rey, CA 90292

## Abstract

With the desire to apply the Dempster-Shafer theory to complex real world problems where the evidential strength is often imprecise and vague, several attempts have been made to generalize the theory. However, the important concept in the D-S theory that the belief and plausibility functions are lower and upper probabilities is no longer preserved in these generalizations. In this paper, we describe a generalized theory of evidence where the degree of belief in a fuzzy set is obtained by minimizing the probability of the fuzzy set under the constraints imposed by a basic probability assignment. To formulate the probabilistic constraint of a fuzzy focal element, we decompose it into a set of consonant non-fuzzy focal elements. By generalizing the compatibility relation to a possibility theory, we are able to justify our generalization to Dempster's rule based on possibility distribution. Our generalization not only extends the application of the D-S theory but also illustrates a way that probability theory and fuzzy set theory can be combined to deal with different kinds of uncertain information in AI systems.

## 1. Introduction

The Dempster-Shafer (D-S) theory of evidence has attracted much attention in AI community in recent years for it suggests a coherent approach, which is sometimes called evidential reasoning, to aggregate evidence bearing on groups of mutually exclusive hypotheses. However, one of the major limitations in its application to plausible reasoning in expert systems is that the theory can not handle evidence bearing on vague concepts.

The knowledge base of an expert system sometimes consists of vague concepts, states that are not well defined, and qualitative descriptions of variables that do not have crisp boundaries. Therefore, an important requirement of a reasoning model for expert systems is its capabilities to manage and aggregate evidence bearing on vague concepts and inexact hypotheses. Although the D-S theory has been extended to deal with imprecise evidential strengths [Zadeh 79, Yager 82, Ishizuka 82, Ogawa 85], no previous extensions have been able to maintain the important principle that the belief and the plausibility functions are lower and upper probabilities.

In this paper, we extend the D-S theory for expressing and combining imprecise evidential strengths in a way such that degrees of belief and degrees of plausibility are still the lower and the upper probabilities constrained by the basic probability assignment. To achieve this, we first generalize the compatibility relation (i.e., multi-valued mapping) in the D-S theory to a joint possibility distribution that captures the degrees of compatibility. In order to express the way a fuzzy focal element constrains the underlying probability distribution, we decompose a fuzzy focal element into a set of consonant non-fuzzy focal elements using the resolution principle in fuzzy set theory. We also generalize Dempster's rule for combining independent sources of inexact evidential strengths. Our extension is justified by employing the noninteractive assumption in possibility theory. Finally, we discuss the similarity and the difference between consonant support functions and possibility distributions.

[1]This article is partially based on the author's Ph.D. thesis at the University of California, Berkeley, which was supported by National Science Foundation Grant DCR-8513139. Part of the research described in the paper was done at USC/Information Sciences Institute and was supported by DARPA under contract No. MDA903-86-C-0178. Views and conclusions contained in this paper are those of the authors, and should not be interpreted as representing the official opinion or policy of the sponsoring agencies.

## 2. Basics of the Dempster-Shafer Theory

We briefly review the basics of the Dempster-Shafer theory in this section [Dempster 67, Shafer 76]. A compatibility relation C between two spaces, S and T, characterizes possibilistic relationships between the elements of two spaces. An element s of S is compatible with an element t of T if it's possible that s is an answer to S and t is an answer to T at the same time [Shafer 84], and the *granule* of s is the set of all elements in T that are compatible with s, i.e., $G(s) = \{ t \mid t \in T, sCt \}$.

Given a probability distribution of space S and a compatibility relation between S and T, a basic probability assignment (bpa) of space T, denoted by $m: 2^T \to [0, 1]$, is induced:

$$m(A) = \frac{\sum_{G(s_i) = A} p(s_i)}{1 - \sum_{G(s_i) = \emptyset} p(s_i)} \tag{1}$$

where the subset A is also called a *focal element*.

The probability distribution of space T, which is referred to as *the frame of discernment*, is constrained by the basic probability assignment. The lower probability and the upper probability of a set B subject to those constraints are called B's belief measures, denoted as Bel(B), and B's plausibility measures, denoted as Pls(B), respectively. These two quantities are obtained from the bpa as follows:

$$Bel(B) = \sum_{A \subseteq B} m(A) \qquad Pls(B) = \sum_{A \cap B \neq \emptyset} m(A). \tag{2}$$

Hence, the belief interval [Bel(B), Pls(B)] is the range of B's probability.

If $m_1$ and $m_2$ are two bpa's induced by two independent evidential sources, the combined bpa is calculated according to Dempster's rule of combination:

$$m_1 \oplus m_2(C) = \frac{\sum_{A_i \cap B_j = C} m_1(A_i) \, m_2(B_j)}{1 - \sum_{A_i \cap B_j = \emptyset} m_1(A_i) \, m_2(B_j)}$$

The basic combining steps that result in Dempster's rule are discussed in Section 4.5.

## 3. Previous Work

Zadeh was the first to generalize the Dempster-Shafer theory to fuzzy sets based on his work on the concept of information granularity and the theory of possibility [Zadeh 79, Zadeh 81]. A possibility distribution, denoted by $\Pi$, is a fuzzy restriction which acts as an elastic constraint on the values of a variable [Zadeh 78, Dubois 88]. Zadeh first generalized the granule of a D-S compatibility relation to a conditional possibility distribution. Then he defined the *expected certainty*, denoted by $EC(B)$, and the *expected possibility*, denoted by $E\Pi(B)$, as a generalization of D-S belief and plausibility functions:

$$E\Pi(B) = \sum_i m(A_i) \sup(B \cap A_i),$$

$$EC(B) = \sum_i m(A_i) \inf(A_i => B) = 1 - E\Pi(B^c)$$

383

where $A_i$ denotes fuzzy focal elements induced from conditional possibility distributions, $\sup(B \cap A_i)$ measures the degree that B intersects with $A_i$, and $\inf(A_i \Rightarrow B)$ measures the degree $A_i$ is included in B. It is easy to verify that the expected possibility and the expected certainty reduce to D-S belief and plausibility measuures when all $A_i$ and $B$ are crisp sets.

Following Zadeh's work, Ishizuka, Yager, and Ogawa have extended the D-S theory to fuzzy sets in slightly different ways [Ishizuka 82, Yager 82, Ogawa 85]. They all extend D-S's belief function by defining a measure of inclusion $I(A \subset B)$, the degree to which set A is included in set B, and by using the following formula similar to Zadeh's expected certainty EC(B):

$$Bel(B) = \sum_{A_i} I(A \subset B)\, m(A_i).$$

Their definitions of the measures of inclusion are listed below.

Ishizuka: $\quad I(A \subset B) = \dfrac{\min_x [1,\, 1+(\mu_B(x)-\mu_A(x))]}{\max_x \mu_A(x)}$

Yager: $\quad I(A \subset B) = \min_x [\mu_{\overline{A}}(x) \vee \mu_B(x)]$

Ogawa: $\quad I(A \subset B) = \dfrac{\sum_i \min[\mu_A(x_i),\, \mu_B(x_i)]}{\sum \mu_B(x_i)}$

Ishizuka and Yager arrive at different inclusion measures by using different implication operators in fuzzy set theory. Ogawa uses relative sigma counts to compute the degree of inclusion.

In order to combine two mass distributions with fuzzy focal elements, Ishizuka extended the Dempster's rule by taking into account the degree of intersection of two sets, J(A,B).

$$m_1 \oplus m_2(C) = \dfrac{\sum_{A_i \cap B_j = C} J(A_i,B_j)\, m_1(A_i)\, m_2(B_j)}{1 - \sum_{i,j} (1 - J(A_i,B_j))\, m_1(A_i)\, m_2(B_j)} \qquad (3)$$

$$\text{where} \quad J(A,B) = \dfrac{\max_x [\mu_{A \cap B}(x)]}{\min[\max_x \mu_A(x),\, \max_x \mu_B(x)]}$$

There are four problems with these extensions. (1) The belief functions sometimes are not sensative to significant change of focal elements because degree of inclusions are determined by certain "critical" points due to the use of "min" and "max" operators. (2) The definitions of "fuzzy inclusion operator" is not unique. Consequently, it is difficult to choose the most appropriate definition for a given application. (3) Although expected possibility and expected certainty (or, equivalently, expected necessity) degenerates to Dempster's lower and upper probabilities in the case of crisp sets, it is not clear that this is a "necessary" extension. (4) The generalized formula for combining evidence is not well justified.

## 4. Our Approach

Instead of directly modifying the formulas in the D-S theory, we generalize the most primitive constructs of the theory and derive other extensions to the theory from these generalizations. Three primitive constructs that have been generalized are the compatibility relation, the objective function for calculating the belief function, and the probabilistic constraints imposed by focal elements. From these generalized basic components, we derive the belief function, the plausibility function, and the rule of combination for our generalized framework.

### 4.1. Generalizing the Compatibility Relation to a Possibility Distribution

The compatibility relation in the D-S theory can only record whether having s and t as answers to S and T respectively is completely possible (i.e., (s,t) is in the relation C) or completely impossible (i.e., (s,t) is not in the relation). In general, however, the possibility that both s and t are answers to S and T is a matter of degree. To cope with this, we generalize Shafer's compatibility relation to a fuzzy relation that records joint possibility distribution of the spaces S and T.

**Definition 1:** A generalized compatibility relation between the spaces S and T is a fuzzy relation $C:2^{S\times T} \to [0,1]$ that represents the joint possibility distribution of the two spaces, i.e., $C(s,t) = \Pi_{X,Y}(s,t)$, where X and Y are variables that take values from the space S and the space T respectively.

Shafer's compatibility relation is a special case of our fuzzy relation where possibility measures are either zeros or ones.

In fuzzy set theory, if the relationship of two variables X and Y are characterized by a fuzzy relation R and the values of variable X is A, the values of variable Y can be induced using the *composition Operation* defined as

$$\mu_{A \circ R}(y) = \max_x \{ \min [ \mu_A(x), \mu_R(x,y) ] \}.$$

So, given a generalized compatibility relation $C:2^{S\times T} \to [0,1]$, the granule of an element s of S is defined to be the composition of the singleton $\{s\}$ and C, i.e.,

$$G(s) = \{s\} \circ C = \Pi_{(Y|X=s)}.$$

Hence, we generalize granules to conditional possibility distributions just like Zadeh did; however, our approach is more general than Zadeh's approach because we go one step further to generalize compatibility relations to join possibility distributions. As we will see in Section 4.5, the generalized compatibility relation is important for justifying our generalization of Dempster's rule.

Given a probability distribution of the space S and a joint possibility distribution between space S and space T such that the granules of S's elements are normal fuzzy subsets,[2] a basic probability assignment (bpa) m to T is induced using equation 1. Adopting the terminology of the D-S theory, we call a fuzzy subset of T with nonzero basic probability a *fuzzy focal element*. A *fuzzy basic probability assignment* (bpa) is a bpa that has at least one fuzzy focal element.

### 4.2. Belief Functions of Fuzzy Sets Induced by Non-fuzzy Bpa's

Pls(B) and Bel(B) are upper and lower probabilities of a set B under the probabilistic constraints imposed by a basic probability assignment. Therefore, Bel(B) can be viewed as the optimal solution of the following linear programming problem:

LP1: Minimize $\sum_{x_i \in B} \sum_j m(x_i : A_j)$ subject to the following constraints:

$$m(x_i : A_j) \geq 0, \qquad i = 1, ..., n; j = 1, ..., l. \tag{4}$$
$$m(x_i : A_j) = 0, \qquad \text{for all } x_i \notin A_j \tag{5}$$
$$\sum_i m(x_i : A_j) = m(A_j) \qquad j = 1, ..., l. \tag{6}$$

Further discussion on formulating the D-S belief function as an optimization problem can be found in [Yen 86]. Since the distribution of each focals' masses do not interfere with each other, we can partition these linear programming problems into subproblems, each one of which concerns the allocation of the mass of a focal element. The optimal solutions of these subproblems are denoted as $m_*(B{:}A_j)$ and $m^*(B{:}A_j)$. Adding the optimal solutions of subproblems, we get B's belief measure and plausibility measure, i.e.,

$$Bel(B) = \sum_{A_j \subseteq T} m_*(B{:}A_j), \quad \text{and} \quad Pls(B) = \sum_{A_j \subseteq T} m^*(B{:}A_j). \tag{7}$$

---

[2] A fuzzy subset A is normal if $\sup_x \mu_A(x) = 1$. The assumption that all focal elements are normal is further discussed in Section 4.5.2.

To compute the belief function of a fuzzy set B induced by a non-fuzzy bpa, we modify the objective functions to account for B's membership function:

Minimize $\sum \sum m(x_i : A_j) \times \mu_B(x_i)$

The optimal solutions of these modified minimization (maximization) problems can be obtained by assigning all the mass of a focal A to an element of A that has the lowest (highest) membership degree in B, i.e.,

$$m_*(B{:}A) = m(A) \times \inf_{x \in A} \mu_B(x), \quad \text{and} \quad m^*(B{:}A) = m(A) \times \sup_{x \in A} \mu_B(x). \tag{8}$$

### 4.3. Decomposing Fuzzy Focal Elements

To deal with fuzzy focal elements, we decompose them into nonfuzzy focal elements. An $\alpha$-level set of A, a fuzzy subset of T, is a crisp set denoted by $A_\alpha$ which comprises all elements of T whose grade of membership in A is greater than or equal to $\alpha$, i.e., $A_\alpha = \{x | \mu_A(x) \geq \alpha\}$. A fuzzy set A may be decomposed into its level-sets through the *resolution identity* [Zadeh 75] $A = \sum_\alpha \alpha A_\alpha$ where the summation denotes the set union operation and $\alpha A_\alpha$ denotes a fuzzy set with a two-valued membership function defined by $\mu_{\alpha A_\alpha}(x) = \alpha$ for $x \in A_\alpha$ and $\mu_{\alpha A_\alpha}(x) = 0$ otherwise.

In order to decompose a fuzzy focal element, we also need to decompose the focal's basic probability and distribute it among the focal's level-sets. We define the decomposition of a fuzzy focal element below.

**Definition 2:** The decomposition of a fuzzy focal element A is a a collection of nonfuzzy subsets such that (1) they are A's $\alpha$-level sets that form a resolution identity, and (2) their basic probabilities are $m(A_{\alpha_i}) = (\alpha_i - \alpha_{i-1}) \times m(A)$, where $i = 1, 2, \ldots, n$, $\alpha_0 = 0$, and $\alpha_n = 1$.

When the focal element is a crisp set, its decomposition is the focal itself because the decomposition contains only one level set, which corresponds to the membership degree "one". It is obvious that the decomposition of a fuzzy focal element form a set of *consonant focals*. The relationship between the decomposition of a fuzzy focal element and Shafer's consonant focals is discussed in Section 4.6.

The probabilistic constraint of a fuzzy focal is defined to be that of its decomposition, which is a set of nonfuzzy focals. Since we already know how to deal with nonfuzzy focals, decomposing a fuzzy focal into nonfuzzy ones allow us to calculate the belief functions that are constrainted by the fuzzy focals.

**Definition 3:** The probability mass that a fuzzy focal A contributes to the belief (and plausibility) of a fuzzy subset B is the total contribution of A's decomposition to B's belief (and plausibility), i.e.,

$$m^*(B:A) = \sum_\alpha m^*(B:A_\alpha), \quad \text{and} \quad m_*(B:A) = \sum_\alpha m_*(B:A_\alpha) \tag{9}$$

### 4.4. Computing the Belief Function

Based on generalizing the objective function and decomposing fuzzy focal elements, we are able to derive the following formula for computing the belief function and the plausibility function.

$$Bel(B) = \sum_A m(A) \sum_{\alpha_i} [\alpha_i - \alpha_{i-1}] \times \inf_{x \in A_{\alpha_i}} \mu_B(x) \tag{10}$$

$$Pls(B) = \sum_A m(A) \sum_{\alpha_i} [\alpha_i - \alpha_{i-1}] \times \sup_{x \in A_{\alpha_i}} \mu_B(x) \tag{11}$$

It is also trivial to show that the dervied formula preserve the following important property of the D-S theory: *The belief of a (fuzzy) set is the difference of one and the plausibility of the set's complement.*

### 4.4.1. An Example

The following example illustrates how one applies the formula described in the previous section for computing the belief function. Suppose the frame of discernment is the set of integers between 1 and 10. A fuzzy basic probability assignment consists of two focal elements A and C:

A = {0.25/1, 0.5/2, 0.75/3, 1/4, 1/5, 0.75/6, 0.5/7, 0.25/8},   C = {0.5/5, 1/6, 0.8/7, 0.4/8},

where each member of the list is in the form of $\mu_A(x_i)/x_i$. We are interested in the degree of belief and the degree of plausibility of the fuzzy subset B:

B = {0.5/2, 1/3, 1/4, 1/5, 0.9/6, 0.6/7, 0.3/8}.

The decomposition of fuzzy focal B consists of four nonfuzzy focals.

$C_{0.4} = \{5, 6, 7, 8\}$ with mass $0.4 \times m(B)$, $C_{0.5} = \{5, 6, 7\}$ with mass $0.1 \times m(B)$
$C_{0.8} = \{6, 7\}$    with mass $0.3 \times m(B)$, $C_1 = \{6\}$    with mass $0.2 \times m(B)$

Similarly, fuzzy focal A can be decomposed into its level sets $A_{0.25}$, $A_{0.5}$, $A_{0.75}$, and $A_1$. Let us denote $\inf_{x \in A_{\alpha_i}} \mu_B(x)$ as $f_{B,A}(\alpha_i)$. So, we have

$m_*(B:A) = m(A) \times [\ 0.25 \times f_{B,A}(0.25) + 0.25 \times f_{B,A}(0.5) + 0.25 \times f_{B,A}(0.75) + 0.25 \times f_{B,A}(1)\ ]$
$= m(A) \times [\ 0.25 \times 0 + 0.25 \times 0.5 + 0.25 \times 0.9 + 0.25 \times 1\ ] = 0.6 \times m(A)$.

Similarly, we get $m_*(B:C) = 0.54 \times m(C)$. Thus, we have $Bel(B) = 0.6\,m(A) + 0.54\,m(C)$. We can also compute the plausibility of B in a similar way: $Pls(B) = m(A) + 0.86\,m(C)$.

## 4.5. Generalizing Dempster's Rule of Combination

Dempster's rule combine the effect of two independent evidential sources, denoted as R and S, on the probability distribution of a hypothesis space, denoted as T. The rule can be viewed as a result of three steps: (1) *Combination of the compatibility relations*: A combined compatibility relation between the product space $R \times S$ and T can be constructed as folloed:

$r\,C\,t$ and $s\,C\,t$ ===> $[r,s]\,C\,t$

where r, s, t, and [r, s] denote elements of R, S, T, and $R \times S$ respectively. (2) *Computing joint probability distributions of the two evidential sources based on the independence assumption*: $P([r,s]) = P(r) \times P(s)$. (3) *Normalizing the combined basic probability assignment to discard probability mass assigned to the empty set*.

Two generalizations have to be made to Dempster's rule before it can be used to combine fuzzy bpa's in our generalized framework. (1) The first step above has to be extended to allow the combination of fuzzy compatibility relations. (2) The normalization step needs to consider subnormal fuzzy focal elements that result from combining fuzzy compatibility relations.

### 4.5.1. Combination of fuzzy compatibility relations

A compatibility relation in our generalized D-S theory, as discussed in Section 4.1, is a joint possibility distribution. Thus, we have $C(r,t) = \Pi_{X,Z}(r,t)$ and $C(s,t) = \Pi_{Y,Z}(s,t)$ where X, Y, and Z are variables that take values from the spaces R, S, and T respectively. Let W be a variable that takes values from the space $R \times S$, the combined fuzzy compatibility relation can be expressed as $C([r,s], t) = \Pi_{W,Z}([r,s], t) = \Pi_{X,Y,Z}(r, s, t)$. By employing the assumption that the variables Y, Z and X, Z are *noninteractive*, a concept analogous to the independence of random variables, we obtain the joint possibility distribution from the two marginal possibility distributions: $\Pi_{X,Y,Z}(r, s, t) = \Pi_{Y,Z}(s, t) \wedge \Pi_{X,Z}(r, t)$. Thus, the combined fuzzy compatibility relation can be obtained from the compatibility relations of two evidential sources: $C([r,s], t) = C(r, t) \wedge C(s, t)$. For a fixed pair of r and s, applying the equation above to all possible elements in T gives us the following relationships between granules:
$$G([r, s]) = G(r) \cap G(s)$$
where $\cap$ denotes fuzzy intersection operator.

### 4.5.2. Normalizing Subnormal Fuzzy Focal Elements

An important assumption of our work is that **all focal elements are normal**. We avoid subnormal fuzzy focal elements because they assign probability mass to the empty set. The probability mass assigned to the empty set by a subnormal fuzzy focal A is $[1 - \max_x \mu_A(x)] \times m(A)$.

Although we have assumed that the focal elements of fuzzy bpa's are all normal, the intersections of focals, however, may be subnormal. Hence, the combination of fuzzy bpa's should deal with the normalization of subnormal fuzzy focal elements. To do this, we need to normalize the two components of a fuzzy focal element: the focal itself, which is a subnormal fuzzy set, and the probability mass assigned to the focal. It's straight forward to normalize the focal. Suppose A is a subnormal fuzzy set characterized by the membership function $\mu_A(x)$. A's normalized set, denoted as $\overline{A}$, is characterized by the membership function $\mu_{\overline{A}}(x) = \mu_A(x)/\max_x \mu_A(x)$.

The criteria for normalizing the probability mass of a subnormal focal is that the probabilistic constraints imposed by the subnormal focal should be preserved after the normalization. Since we use the decomposition of a focal to represent its probabilistic constraint, this means that the probability mass assigned to a decomposed focal should not be changed by the normalization process. Since the $\alpha_i$ cut of the subnormal focal becomes the $k\alpha_i$ cut of the normalized focal, where k is the normalization factor $k = 1/\max_x \mu_A(x)$, the probability mass assigned to them should be the same, i.e.,

$$m(A_{\alpha_i}) = m(\overline{A}_{k\alpha_i}). \tag{12}$$

From this condition and the definition of fuzzy focal's decompositions, we can derive the following relationship between $m(\overline{A})$ and $m(A)$: $m(\overline{A}) = m(A)/k$. Hence, mass of the normalized focal is reduced at the same rate its membership function has been scaled up. The remaining mass $(1 - 1/k) m(A)$ is the amount assigned to the empty set by the subnormal fuzzy focal and, hence, should be part of the normalization factor in the generalized Dempster's rule.

### 4.5.3. A Generalized Dempster's Rule

If $m_1$ and $m_2$ are two fuzzy bpa's induced by two independent evidential sources, the combined bpa is calculated according to the generalized Dempster's rule of combination:

$$m1 \oplus m2(C) = \frac{\sum_{(\overline{A} \cap \overline{B}) = C} \max_{x_i} \mu_{A \cap B}(x_i) \, m1(A) \, m2(B)}{1 - \sum_{A, B} (1 - \max_{x_i} \mu_{A \cap B}(x_i)) \, m1(A) \, m2(B)} \tag{13}$$

Our extension generalizes the notion of *conflicting evidence* in the D-S theory to that of *partially conflicting evidence*. In original Dempster's rule, two pieces of evidence are either in conflict (i.e., the intersection of their focals is empty) or not in conflict at all (i.e., the intersection of their focals is not empty). In our generalized combining rule, two pieces of evidence are partially in conflict if the intersection of their focals is subnormal. The degree of conflict is determined by the peak (i.e., the maximum value) of the focal's membership function. The case of peak being zero corresponds to the case of total conflict in the D-S theory.

When both set A and set B are normal, Ishizuka's degree of intersection becomes

$J(A,B) = \max_{x_i} \mu_{A \cap B}(x_i)$. His extension, formulated in equation 3, reduces to ours. However, unlike our approach, Ishizuka's extension is not justified using the possibility theory and the probability theory.

Although a fuzzy focal can be viewed as a set of consonant crisp focals in calculating belief functions, it differs from consonant focals in the evidence it represents. A fuzzy focal represents **one piece of evidence**, while a set of consonant focals represent **several pieces of evidence from one evidential source**, which is called the *inferential evidence* in [Shafer 76].

### 4.6. Relationships to Consonant Support Functions

Several authors have discussed the similarity between possibility distribution and the plausibility function when the focal elements are nested, i.e., they can be arranged in order so that each is contained in the following one [Shafer 84]. In this section, we first explain why the plausibility of consonant focals exhibit similar properties as possibility distribution, then we discuss their differences.

The plausibility function induced by a set of consonant focals is equivalent to that induced by a fuzzy set that is composed from the consonant focals. Suppose $A_1$, $A_2$, ..., $A_n$ are consonant focals such that $A_1 \subset A_2 \subset A_3 ... \subset A_n$. Then we can compose a fuzzy focal element A by treating $A_k$ as a level set of A whose membership degree (i.e., $\alpha$ value) is $\sum_{i=1}^{k} m(A_i)$. It is straight forward to see that the decomposition of A is exactly the set of given consonant focals. Therefore, the plausibility function induced by the consonant focals is the same as that induced by the constructed fuzzy focal.

To show the relationship between consonant plausibility function and possibility distribution, we first prove that the highest basic probability that a fuzzy focal element can assign to an element is proportional to the degree that the element belongs to the fuzzy focal.

**Theorem 4:** Let m be a bpa that assigns nonzero mass to a fuzzy subset A of the frame of discernment T, the maximum mass that can be allocated to an element t of T is proportional to the membership of t in A, i.e., $\mu_A(x) \times a$, i.e.,

$$m^*(\{t\} : A) = m(A) \times \mu_A(t). \tag{14}$$

**Proof:** Let $\mu_A(t) = \alpha_k$. All the decomposed focals whose $\alpha$ values are smaller than $\alpha_k$ could contribute all of their masses to t. However, the decomposed focals whose $\alpha$ values are higher than the membership degree of t can not assign their masses to t. Therefore, we have

$$m^*(\{t\} : A) = m(A) \times \sum_{i=1}^{k}(\alpha_i - \alpha_{i-1}) = m(A) \times \alpha_k = m(A) \times \mu_A(t)$$

It follows from the theorem that when a bpa consists of only one fuzzy focal element A, i.e., $m(A) = 1$, the plausibility of an element t becomes the membership degree of t in fuzzy set A, which characterizes a conditional possibility distribution, i.e., $Pls(\{t\}) = \mu_A(t)$. Since consonant focal elements can be used to construct a fuzzy focal element, the equation above explains why the plausibility of consonant focals exhibit similar properties as that of possibility measures.

Although consonant support function behaves like possibility distribution, it differs from possibility distributions in two major ways. (1) Consonant support functions are more restrictive in the kinds of evidence they can represent. More specifically, they are not appropriate for representing multiple fuzzy focal elements that are induced from a joint possibility distribution. (2) Combining consonant support function do not always yield another consonant support function. However, combining possibility distributions in our framework always result in another possibility distribution.

## 5. Conclusions

We have described a generalization of the Dempster-Shafer theory to fuzzy sets. Rather than generalizing the formula for computing belief function, we generalize the basic constructs of the D-S theory: the compatibility relations, the objective functions of the optimization problem for calculating belief functions, and the probabilistic constraints imposed by focal elements. As a result, we can compute the lower probability (i.e., the belief function) directly from these generalized constructs. Moreover, by employing the noninteractive assumption in possibility theory, we modified Dempster's rule to combine evidence that may be partially in conflict.

Our approach offers several advantages over previous work. First, the semantics of the D-S theory is maintained. Belief functions are lower probabilities in our extension. Second, we avoid the problem of "choosing the right inclusion operators" faced by all previous approaches. Third, the belief function are determined by the shape of the whole focal elements, not just a critical point. Any change of focal elements directly affects its probabilistic constraints, which in turn affects the belief function. Finally, through our generalization, we have shown that the D-S theory could serve as a bridge that brings together probability theory and fuzzy set theory into a hybrid approach to reasoning under uncertainty and inexactness.

## Acknowledgement

The author would like to thank Prof. Lotfi Zadeh for suggesting viewing consonant focals as a decomposition of a fuzzy focal element and for his comments on an earlier draft of the paper.